\documentclass[10pt,twocolumn,letterpaper]{article}
\usepackage[pagenumbers]{cvpr}
\usepackage{amsthm}
\usepackage{algorithm}
\usepackage{algpseudocode}
\usepackage{placeins}
\usepackage{needspace}
\usepackage{multirow}
\usepackage{tabularx}
\definecolor{cvprblue}{rgb}{0.21,0.49,0.74}
\usepackage[pagebackref,breaklinks,colorlinks,allcolors=cvprblue]{hyperref}
\usepackage[capitalize]{cleveref}
\AtBeginDocument{%
  \Crefname{section}{Sec.}{Secs.}%
  \crefname{section}{Section}{Sections}%
  \Crefname{table}{Tab.}{Tabs.}%
  \crefname{table}{Table}{Tables}%
  \Crefname{figure}{Figure}{Figures}%
  \crefname{figure}{Fig.}{Figs.}%
  \crefname{theorem}{Theorem}{Theorems}%
  \crefname{proposition}{Proposition}{Propositions}%
  \crefname{corollary}{Corollary}{Corollaries}}
\def\confName{CVPR}
\def\confYear{2027}
\providecommand{\Description}[1]{}
\newcommand{\codename}{VoteAlign}
\newcommand{\FPR}{\mathrm{FPR}}
\newcommand{\TPR}{\mathrm{TPR}_{\mathrm{succ}}}

\newcommand{\ind}{\mathbf{1}}
\title{{\codename}: SVM-Confirmed Mean Fusion for Low-False-Positive Adversarial Detection}
\author{Jie Wan \qquad Ziqi Yang\thanks{Corresponding author: Ziqi Yang (\texttt{yangziqi@zju.edu.cn}).}\\
Zhejiang University\\
{\tt\small wanjie@zju.edu.cn \quad yangziqi@zju.edu.cn}}
\begin{document}
\maketitle
\begin{abstract}
Adversarial attacks pose a persistent threat to vision--language models.
Many adversarial detectors flag images by thresholding scores based on internal features, changes in model responses to image transformations, or image--text similarity.
When normal images score as high as attacks, a threshold that catches those attacks also flags the normal images.
We propose {\codename} to improve detection while reducing false alarms.
It averages head scores based on several CLIP layers and sets the threshold on normal images.
A fixed linear support vector machine uses the same encoding to check alarms and can reject high-scoring normal images at that threshold.
Our theoretical proofs give conditions for ranking losses to improve mean-branch detection at a fixed false-positive rate along smooth, stable fitting paths.
Finite-sample guarantees require independent normal calibration and attack audits.
On shared held-out PGD-40 test inputs, {\codename} detects 99.97\% of successful attacks versus 95.56\% for the strongest baseline, with 27 false alarms versus 89 among 100,000 normal images.

\end{abstract}
\section{Introduction}
\label{sec:introduction}
Vision--language models such as CLIP~\cite{radford2021clip} use class names in text prompts to recognize images without retraining the image encoder.
Small adversarial perturbations can change a classifier's prediction while leaving the visible content largely unchanged~\cite{szegedy2014intriguing,goodfellow2015explaining,wan2025btuap}.
An added detector can flag these inputs, but frequent false alarms on normal images can make it hard to use even when it catches most attacks.
We study detection for frozen CLIP at a low false-positive rate (FPR), where score construction and threshold selection must work together~\cite{tong2018np}.

Existing adversarial detectors construct scores from original-image features, responses to input transformations, or image--text similarities.
Feature-based methods train binary detectors on classifier activations~\cite{metzen2017detecting} or score feature distributions~\cite{lee2018mahalanobis}.
Feature Squeezing~\cite{xu2018squeezing}, CSR~\cite{nie2026csr}, and TTP~\cite{li2026ttp} use transformation-based signals, while A4D~\cite{krakover2026a4d} uses text descriptions of attack artifacts.
Detectors also aggregate evidence across layers: the LID variant of GeoDetect combines layer scores using logistic regression~\cite{hasanebrahimi2026geodetect}, and SAEgis averages layer-wise activation statistics~\cite{wang2026saegis}.
For threshold selection, SAEgis uses an empirical quantile of normal scores~\cite{wang2026saegis}, while Neyman--Pearson calibration selects a normal-score order statistic to provide finite-sample FPR control under independent calibration~\cite{tong2018np}.

However, high detection rates with these methods often come at the cost of elevated false-positive rates.
Controlling false positives while preserving detection performance introduces three connected challenges.
First, low-FPR operation requires separation specifically in the upper tail of normal scores.
At a stringent FPR target, the threshold is determined by a small fraction of high-scoring normal inputs, so good overall clean--attack discrimination need not preserve detection at that threshold.
Learning objectives must therefore distinguish attacks from these difficult normal cases.
Second, reducing false alarms requires selective use of heterogeneous detection evidence.
Different signals can vary in scale and strength, affecting their relative influence on a combined score.
If that score ranks a normal input above an attack, increasing its threshold to reject the normal input also rejects the attack.
The challenge is to combine continuous evidence and use discriminative information beyond this ordering to selectively suppress false alarms, while accounting for any lost detections.
Third, the target FPR must be supported by population-level error control from finite data.
A small empirical false-positive rate does not by itself establish that a target population FPR is satisfied, and controlling FPR alone does not establish a high TPR.
The challenge is to calibrate the deployed rule under explicit sampling assumptions and quantify its detection performance and uncertainty.

We propose {\codename}, an adversarial detector designed to substantially reduce false alarms while maintaining high detection rates.
The method combines three independently trained detection heads with a fixed linear SVM, all using features from the CLIP pass used for classification.
First, we introduce three detection heads with classification and head-specific ranking objectives to distinguish attacks from high-scoring normal inputs.
The heads use features from several CLIP blocks or the final block alone, with ranking terms in selected heads targeting difficult normal images and clean--attack pairs.
Second, we align the continuous head scores using normal-data reference statistics and average them equally to retain score strength while accounting for scale differences.
Third, we introduce a fixed linear SVM confirmation rule on the original block features to selectively reject false alarms from the mean alarm set.
Last, we apply normal-data Neyman--Pearson calibration to the mean branch to control the final detector's FPR under the calibration assumptions.
We freeze the heads, score alignment, and SVM before calibration, and flag an input only when its mean score exceeds the calibrated threshold and the SVM decision is positive.
Overall, we analyze when ranking losses can improve mean-branch detection at fixed FPR along smooth, stable fitting paths.
We characterize the SVM confirmation tradeoff and derive finite-sample FPR and TPR guarantees under independent normal calibration and attack auditing in the appendix.

\paragraph{Contributions.}
\begin{itemize}
\item \textbf{Method design.}
  We propose {\codename} to reduce false alarms while retaining high detection rates.
  Our design combines classification, hard-normal ranking, and paired-ranking objectives with aligned mean fusion, fixed SVM confirmation, and normal-data Neyman--Pearson calibration.
\item \textbf{Theoretical analysis.}
  We establish conditions for ranking losses to improve mean-branch detection at fixed FPR along smooth, stable fitting paths.
  Within a fixed candidate set, we bound the TPR advantage of any gate with no higher FPR than the SVM gate.
  For the final detection rule, we derive finite-sample FPR and TPR guarantees under independent normal calibration and attack auditing.
\item \textbf{Experimental validation.}
  On shared held-out PGD-40 test inputs at a common nominal FPR target, {\codename} detects 99.97\% of successful attacks versus 95.56\% for TTP, with 27 false alarms versus 89 among 100,000 normal images.
  Across five attack settings, two backbones, and three target domains, we assess detection and transfer; branch, component, and data-size studies quantify how design choices affect false alarms and missed attacks.
\end{itemize}

\section{Related Work}
\label{sec:related}
\subsection{Adversarial Attacks}
White-box attacks use access to the target model to optimize small input perturbations~\cite{szegedy2014intriguing}.
FGSM~\cite{goodfellow2015explaining} takes one gradient-sign step, and PGD~\cite{madry2018towards} iterates under a perturbation budget.
DeepFool~\cite{moosavi2016deepfool} and Carlini--Wagner attacks~\cite{carlini2017evaluating} seek small changes that cross the classifier's decision boundary.

Black-box attacks use surrogate models or query responses.
Papernot et al.~\cite{papernot2017practical} use target predictions to train a substitute model and generate attacks that transfer.
Dong et al.~\cite{dong2018momentum} use momentum to improve transfer across models.
Query-based methods can use prediction scores~\cite{ilyas2018blackbox} or only class labels.
Boundary Attack~\cite{brendel2018decision} refines an adversarial starting point with label queries.
Gradient-estimation methods use the same feedback to estimate local gradient directions at the boundary and guide the search toward adversarial examples closer to their original inputs~\cite{chen2020hopskipjump,wan2024bounceattack}.

AutoAttack~\cite{croce2020reliable} combines white-box and black-box components for robustness evaluation.
Detector evaluations distinguish attacks that target the classifier from adaptive attacks that also account for the detector~\cite{carlini2017detection,tramer2020adaptive}.

\subsection{Adversarial Detection}
Detection scores can use features of the original image, changes under input transformations, or similarities to text prompts.
Metzen et al.~\cite{metzen2017detecting} train a separate network on classifier features, while Pang et al.~\cite{pang2018robustdetection} train for better separation of normal and adversarial representations.
Other signals include uncertainty and feature density~\cite{feinman2017artifacts}, Mahalanobis distance~\cite{lee2018mahalanobis}, and local intrinsic dimensionality (LID)~\cite{ma2018lid}.
Maximum softmax confidence is a related baseline for detecting misclassified and out-of-distribution inputs~\cite{hendrycks2017baseline}.

Other scores use input transformations or text prompts.
Feature Squeezing~\cite{xu2018squeezing} uses quantization and smoothing, TTP~\cite{li2026ttp} uses spatial padding, and Defend-CLIP~\cite{defendclipcode2026} compares feature drift across noise levels.
CSR~\cite{nie2026csr} uses changes in features after frequency filtering to guide test-time correction.
We use its spectral discrepancy as a detection score in the baseline comparison.
The effect of filtering depends on the frequency content of the perturbation~\cite{yin2019fourier,guo2020lowfrequency}.
A4D~\cite{krakover2026a4d} scores image similarity to text descriptions of attack artifacts without accessing the target classifier.

Detectors also differ in how they combine evidence across layers.
Deep k-Nearest Neighbors~\cite{papernot2018dknn} checks label support across layers, NIC~\cite{ma2019nic} checks activation and provenance invariants, and Layer Regression~\cite{mumcu2025layer} scores errors in predicting deeper features from earlier ones.
The LID variant of GeoDetect~\cite{hasanebrahimi2026geodetect} combines layer scores using logistic regression.
SAEgis~\cite{wang2026saegis} averages sparse activation statistics across layers.
General ensemble studies distinguish training the members from combining their scores~\cite{dietterich2000ensemble,breiman1996bagging,freund1997boosting,lakshminarayanan2017ensembles} and relate voting performance to member quality and disagreement~\cite{germain2015majority,kuncheva2003diversity}.
Hard-example mining~\cite{shrivastava2016ohem} and focal loss~\cite{lin2017focal} give difficult examples more weight during training.

Threshold calibration sets the operating point of a fixed score.
SAEgis uses an empirical quantile of normal scores.
Neyman--Pearson (NP) calibration uses an independent sample of i.i.d. normal inputs to bound the probability of exceeding a target population FPR~\cite{tong2018np}.
Conformal outlier p-values~\cite{bates2023outliers} offer another way to test whether inputs differ from normal data, and binomial confidence limits~\cite{clopper1934limits} measure uncertainty in observed false-positive rates.

At low FPR, overlap with the upper tail of normal scores can cause missed attacks.
Fusion can be sensitive to score scale, and calibration alone leaves score separation unchanged.
Transformation scores require extra evaluations, and text-based scores depend on prompt choice.
{\codename} intersects a calibrated mean of aligned scores with a fixed feature-space boundary, using one image encoding.
We measure the resulting tradeoff between detection and false alarms at low-FPR operating points.

\section{Threat Model}
\label{sec:threat-model}
We consider untargeted image perturbations against a frozen CLIP classifier under white-box access to that classifier.
Given a clean image $x^c$ with a correct prediction, the attacker seeks a perturbed image $x^a$ that induces an incorrect class prediction while retaining the original ground-truth label.
For pixel values in $[0,1]$, the evaluated perturbations satisfy $\|x^a-x^c\|_\infty\le 3/255$.
The attacker has access to the classifier's architecture and parameters and can compute its outputs and input gradients to modify the image within this budget.
Classifier weights, detector parameters, and calibration statistics are fixed during evaluation.
Attack generation optimizes classifier misclassification, and the resulting images are subsequently scored by each detector.
We evaluate PGD-40, PGD-100, APGD-CE, APGD-DLR, and AutoAttack-standard at the same budget (\cref{sec:experimental-setup}); further details are provided in the appendix.
Detection rates are computed on inputs that the classifier labels correctly before perturbation and incorrectly afterward, using shared attack images and eligibility masks across detectors.
The defender seeks to flag these successful attacks while maintaining a low false-positive rate on normal images.

\section{{\codename}}
\label{sec:method}
\begin{figure*}[t]
\centering
\includegraphics[width=\textwidth]{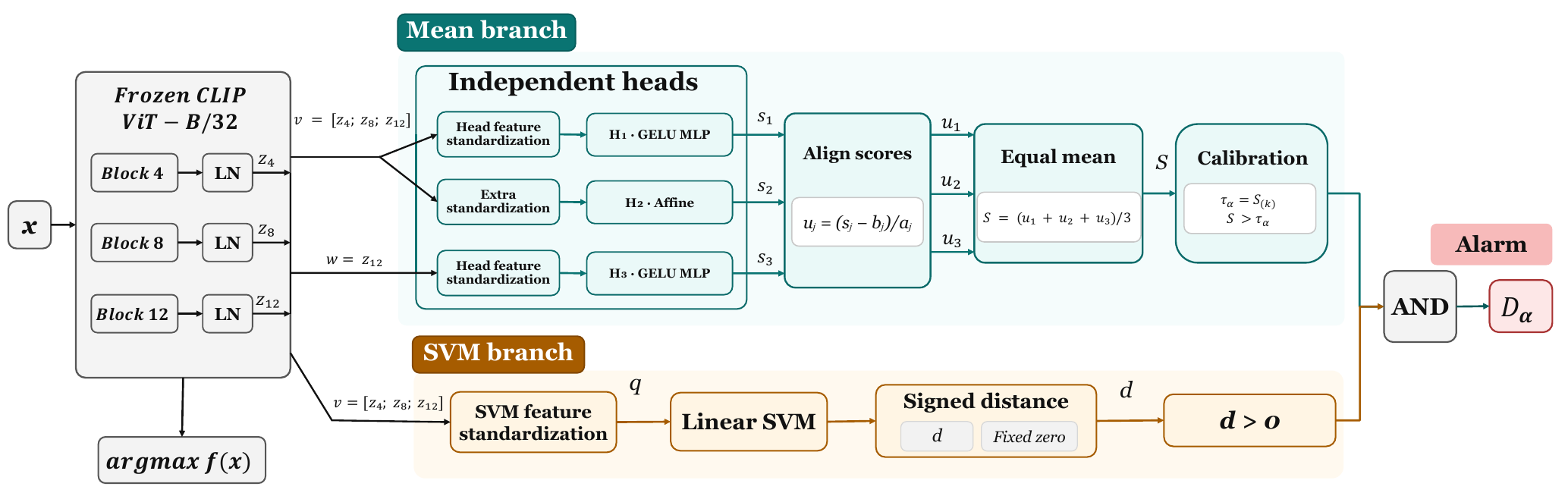}
\caption{Overview of {\codename}.
  One frozen CLIP encoding supplies three independently trained heads and a separately fitted linear SVM.
  The $H_2$ path adds a second feature standardization after the shared one.
  The mean equally weights centered, IQR-scaled head scores.
  Normal-data Neyman--Pearson calibration sets $\tau_\alpha$.
  An alarm requires both $S(x)>\tau_\alpha$ and the fixed SVM condition $d(x)>0$.}
\label{fig:overview}
\Description{The input image passes once through frozen CLIP.
  Blocks 4, 8, and 12 feed two detection heads and the linear SVM; block 12 also feeds a third head.
  Normalized head scores are averaged and compared with a calibrated threshold.
  The final alarm is the logical AND of this mean decision and the SVM's positive-distance decision.
  All fitted models remain frozen at inference.}
\end{figure*}

{\codename} aims to reduce false alarms while maintaining high detection rates for adversarial inputs to a frozen CLIP classifier.
As shown in \cref{fig:overview}, we first encode an image once with frozen CLIP to obtain its class prediction and intermediate block features.
We then use three independently trained heads to score these features, align the scores using fixed normal-data reference statistics, and average them equally.
Finally, we compare the mean score with a threshold set by normal-data Neyman--Pearson calibration and confirm candidate alarms with a fixed linear SVM on the shared block features.
An input is flagged only when its mean score exceeds the calibrated threshold and the SVM decision is positive.
\subsection{Design Motivation}
\label{sec:motivation}
Our starting point is a limitation of thresholding: when a normal image scores above an attack, raising the threshold to remove that false alarm also misses the attack.
At low FPR, these high-scoring normal images determine the operating point, so improving average clean--attack separation alone need not resolve the problem.

This suggests two distinct tasks: improving how inputs are ranked and checking which candidate alarms should be retained.
Ranking objectives can target difficult normal inputs, while the original block features may retain distinctions not captured by a scalar mean.
The clean--adversarial response differences reported by BTUAP~\cite{wan2025btuap} provide further motivation to explore this feature-space check.

\label{sec:svm-motivation}
We therefore align continuous head scores before averaging and confirm mean alarms with a fixed linear SVM on the block features.
Normal-data calibration sets the mean threshold while leaving the confirmation boundary unchanged, giving score learning, alarm screening, and threshold selection separate roles.

\subsection{Shared Features and Independent Heads}
\label{sec:vote-overview}
\label{sec:vote-heads}
To score attack evidence at different feature depths, we use three independent heads that share a frozen CLIP encoding.
For ViT-B/32, let $h_\ell(x)\in\mathbb R^{768}$ be the class token after block $\ell$ and $z_\ell(x)=\operatorname{LN}(h_\ell(x))$ its parameter-free layer normalization.
We form two feature vectors
\begin{equation}
 v(x)=[z_4(x);z_8(x);z_{12}(x)],\qquad w(x)=z_{12}(x).
 \label{eq:features}
\end{equation}

For the heads, we standardize $v$ and $w$ in Eq.~\eqref{eq:features} coordinate by coordinate using normal data.
$H_1$ combines features across blocks with a $2304\rightarrow64\rightarrow1$ GELU MLP on $v$.
$H_2$ scores the same vector with an affine function after a second standardization fitted on its normal training pool.
$H_3$ focuses on the final block with a $768\rightarrow64\rightarrow1$ GELU MLP on $w$.
Each head gives an unbounded score $s_j$, with higher values indicating stronger attack evidence.
The mean branch combines these scores, and a separate SVM uses $v$ with its own standardization to confirm the mean alarms (\cref{sec:vote-decision}).

\subsection{Independent Head Training}
\label{sec:vote-objective}
In order to separate attacks from normal inputs while targeting high-scoring normal images, we train each head with its own objective and select its checkpoint independently.
Each head uses a classification loss, and two heads add ranking losses.
Let $(x_i^c,x_i^a)$ be clean--attack training pairs, $\mathcal B_N$ a sampled normal batch, and $\phi(t)=\log(1+e^t)$.
All heads use
\begin{equation}
 \mathcal L_{C,j}=\mathbb E_i\phi(-s_j(x_i^a))+
 \mathbb E_{x\in\{x_i^c\}\uplus\mathcal B_N}\phi(s_j(x)).
 \label{eq:head-bce}
\end{equation}
We average the attack and normal terms in Eq.~\eqref{eq:head-bce} separately.
The multiset union $\uplus$ retains repeated sampled images.
The paired-ranking and hard-normal losses are
\begin{align}
 \mathcal L_{P,j}&=\mathbb E_i\phi(1+s_j(x_i^c)-s_j(x_i^a)),\notag\\
 \mathcal L_{H,j}&=\mathbb E_{z\in\mathcal H_j,i}
                  \phi(1+s_j(z)-s_j(x_i^a)).
 \label{eq:head-rank}
\end{align}
At each epoch boundary, we set $\mathcal H_j$ to the top-scoring 1\% of the full normal training pool under head $j$, including the clean members of the training pairs.
We hold this set fixed within the epoch.
The hard-normal term in Eq.~\eqref{eq:head-rank} averages all sampled hard-normal and attack combinations.
The three objectives are
\begin{align}
 \mathcal L_1&=\mathcal L_{C,1},\notag\\
 \mathcal L_2&=\mathcal L_{C,2}+0.5\mathcal L_{H,2},\notag\\
 \mathcal L_3&=\mathcal L_{C,3}+0.5\mathcal L_{P,3}+0.5\mathcal L_{H,3}.
 \label{eq:individual}
\end{align}
The classification loss trains each head to separate clean and attack scores.
Hard-normal ranking targets high-scoring normal images in $H_2$ and $H_3$, and $H_3$ also ranks attacks against their clean sources.
With a continuous normal-score CDF, population hard-normal risk bounds the normalized partial-AUC deficit for the same score and tail, as shown in the appendix.
Training uses a 1\% normal tail for each head, while evaluation uses a 0.1\% ROC interval.
Applying the bound to the mean or diagnostic score requires that score's matching-tail risk.
Eq.~\eqref{eq:loss-sensitivity} gives a separate local condition linking the ranking weights to mean-branch detection.

\subsection{Mean Fusion and SVM Confirmation}
\label{sec:vote-decision}
We seek to combine attack evidence and reduce false alarms by aligning and averaging the head scores, then confirming mean alarms with a fixed linear SVM.
For head $j$, we fix the normal-score reference $b_j$ and the normal-score interquartile range (IQR) $a_j$, floored at $10^{-6}$, and compute
\begin{equation}
 u_j(x)=\frac{s_j(x)-b_j}{a_j},\qquad
 S(x)=\frac13\sum_{j=1}^3u_j(x).
 \label{eq:margins}
\end{equation}
The references in Eq.~\eqref{eq:margins} shift $S$ by a constant that calibration absorbs.
The IQRs set the relative scales and can change the ranking.
We fix both statistics and the equal weights before calibrating the fitted detector.

\paragraph{Ranking-loss effect at fixed FPR.}
Let $\xi$ denote head parameters after removing score-preserving directions.
For fixed normal and successful-attack distributions, let $T_\alpha(\xi)$ be the mean branch's TPR after adjusting its population threshold to maintain FPR $\alpha$.
Consider a smooth stationary path $\xi_\lambda$ for $\mathcal J_\lambda=\mathcal J_0+\lambda\mathcal R$, with reduced Hessian $H_\lambda=\nabla_\xi^2\mathcal J_\lambda(\xi_\lambda)\succ0$, differentiable score-boundary probabilities, and positive normal-score density at the threshold.
Here $\mathcal R$ is either $\mathcal L_{H,2}+\mathcal L_{H,3}$ or $\mathcal L_{P,3}$, and $\mathcal J_0$ contains the remaining terms in Eq.~\eqref{eq:individual}.
Under the regularity conditions given in the appendix,
\begin{equation}
 \frac{\mathrm d}{\mathrm d\lambda}T_\alpha(\xi_\lambda)
 =-\bigl(\nabla_\xi T_\alpha\bigr)^{\top}H_\lambda^{-1}\nabla_\xi\mathcal R.
 \label{eq:loss-sensitivity}
\end{equation}
All gradients are evaluated at $\xi_\lambda$.
The TPR gradient includes changes in alignment and the population threshold.
A positive right-hand side gives a local TPR gain as that ranking weight increases at fixed mean-branch FPR.
The fixed-gate intersection has its own boundary condition given in the appendix.

A separate linear SVM~\cite{cortes1995support} checks mean alarms against the block features.
With coordinate means $\mu_N$ and standard deviations $\sigma_N$ from the normal training images, we compute its signed distance as
\begin{equation}
 q(x)=\frac{v(x)-\mu_N}{\sigma_N},\qquad
 d(x)=\frac{\beta^\top q(x)+c}{\|\beta\|_2}.
 \label{eq:svm-distance}
\end{equation}
The standardization in Eq.~\eqref{eq:svm-distance} is coordinatewise.
We fit the SVM with balanced class weights and PGD attacks as the positive class.
Its statistics, parameters, and zero boundary stay fixed during calibration and inference.
With mean threshold $\tau_\alpha$ calibrated on normal images, the deployed rule is
\begin{equation}
 D_\alpha(x)=\ind\{S(x)>\tau_\alpha\}\,\ind\{d(x)>0\}.
 \label{eq:mean-alarm}
\end{equation}
Both tests are strict.
At a fixed mean threshold, the reductions in FPR and TPR equal the probabilities of rejected mean alarms on normal and attack inputs, respectively.
Within the same candidate set, the analysis in the appendix bounds the TPR advantage of any gate with no higher FPR than the SVM gate.
The bound uses squared-hinge excess risk under the evaluation laws.
The empirical loss uses class averages from the training pools.
Recalibration changes only the mean threshold.

\paragraph{Diagnostic ranking.}
\label{sec:ranking-diagnostic}
For the diagnostic low-FPR ROC, we rank images by
\begin{equation}
 R(x)=\min\{S(x)-\tau_{\rm primary},\ d(x)/s_{\rm fixed}\},
 \label{eq:diagnostic-score}
\end{equation}
where $s_{\rm fixed}$ is the IQR of $d$ on development normals, floored at $10^{-6}$.
Threshold zero in Eq.~\eqref{eq:diagnostic-score} reproduces Eq.~\eqref{eq:mean-alarm} at the primary operating point.
The ROC and npAUC describe ranking as a sweep of $R$ moves both boundaries.
For deployment, we calibrate $S$ and keep the SVM boundary at zero.

\begin{algorithm}[t]
\caption{Training, calibration, and inference for {\codename}.}
\label{alg:votealign}
\begin{algorithmic}[1]
\Require Frozen CLIP $f$; normal and paired PGD training data;
development data; normal reference and calibration samples;
target FPR $\alpha$, tolerance $\delta$; query image $x$
\State Cache block features and fit normal feature statistics
\For{$j=1,2,3$}
 \State Train head $j$ with Eq.~\eqref{eq:individual}
 \State Freeze its development-selected checkpoint
\EndFor
\State Fix $b_j$ and $a_j$ on the normal reference sample
\State Fit normal-coordinate statistics and select a linear SVM
\State Freeze its parameters and the boundary $d=0$
\State Score each normal calibration image with $S$ and set $\tau_\alpha$ by Eq.~\eqref{eq:np}
\State Encode $x$ once to obtain logits $f(x)$ and features $v(x),w(x)$
\State Compute $S(x)$ and signed distance $d(x)$
\State \Return $\arg\max f(x)$ and $D_\alpha(x)$ by Eq.~\eqref{eq:mean-alarm}
\end{algorithmic}
\end{algorithm}

\subsection{Mean-Branch Calibration}
\label{sec:vote-calibration}
The calibration step aims to control the false-positive rate by applying the Neyman--Pearson rule to normal scores from the fixed mean branch.
Algorithm~\ref{alg:votealign} summarizes fitting, calibration, and inference.
Sort $n$ normal calibration scores as $S_{(1)}\le\cdots\le S_{(n)}$.
For nominal FPR $\alpha$ and violation tolerance $\delta$, the Neyman--Pearson (NP) rule~\cite{tong2018np} selects
\begin{equation}
 k=\min\!\left\{r:\sum_{i=r}^{n}\binom{n}{i}(1-\alpha)^i
                          \alpha^{n-i}\le\delta\right\},\qquad
 \tau_\alpha=S_{(k)}.
 \label{eq:np}
\end{equation}
A finite threshold requires a rank that meets Eq.~\eqref{eq:np}.
For a fixed score and independent i.i.d.\ target normals, the proof in the appendix shows that the final rule's population FPR exceeds $\alpha$ with probability at most $\delta$ over calibration draws.
The strict score comparison preserves this bound when scores tie.
Every confirmed alarm is a mean alarm, so this result holds for any dependence between branches.
Calibration on source normals gives a source-population guarantee.
For TPR, theoretical bounds for the frozen rule using independent attack audits are given in the appendix.
The source-audit bound additionally requires control of the source-to-target attack shift.
Model selection is completed before final calibration; detailed selection rules are provided in the appendix.

\section{Experiments}
\label{sec:experiments}
We evaluate low-FPR detection, transfer across attacks and domains, and the roles of the two decision branches.

\subsection{Experimental Setup}
\label{sec:experimental-setup}
\paragraph{Data and attacks.}
We use frozen CLIP ViT-B/32~\cite{radford2021clip,dosovitskiy2021vit} on ImageNet~\cite{deng2009imagenet}, with $224\times224$ inputs.
All attacks target the CLIP classifier at $\ell_\infty$ budget $3/255$, also used for training and development.
The primary ImageNet comparisons and component studies share a held-out cohort of 50,000 calibration normals, 100,000 test normals, and 5,000 attack sources.
Identity and original-file hash checks separate these roles from one another and from fitting, model selection, and score-reference estimation.
Each clean--attack pair stays in one split, and all methods in a comparison share the saved attacks and success mask.
The cohort also appears in prior reports, and these analyses reuse its evaluation images.
Data roles and attack settings appear in the appendix.

\paragraph{Fitting and comparison methods.}
The three heads train independently on 8,000 PGD pairs, with checkpoints selected on development data.
After fixing the heads, SVM, and score-reference statistics, we calibrate the mean threshold on independent normals using the NP rule in \cref{sec:vote-calibration} at target FPR $\alpha=0.001$ and tolerance $\delta=0.05$.
The SVM retains its zero boundary, and baselines use the same NP rule and normal calibration images.

We compare CSR~\cite{nie2026csr}, TTP~\cite{li2026ttp}, Feature Squeezing~\cite{xu2018squeezing}, Defend-CLIP~\cite{defendclipcode2026}, a GeoLID adaptation of GeoDetect~\cite{hasanebrahimi2026geodetect}, and two internal residual-feature controls on the same CLIP backend.
GeoLID's B/32 head uses 256 PGD training pairs.
Comparisons against {\codename} at matched pair counts appear in the appendix.
The other baselines fit no PGD-supervised head.
We measure detection before any correction or adaptation.
Full fitting and baseline settings appear in the appendix.

\paragraph{Metrics.}
For normal images $\mathcal B$ and successful classifier attacks $\mathcal A$, TP counts detected attacks and FP counts normal alarms:
\begin{equation}
 \FPR=\frac{\sum_{x\in\mathcal B}D(x)}{|\mathcal B|},\qquad
 \TPR=\frac{\sum_{x\in\mathcal A}D(x)}{|\mathcal A|}.
 \label{eq:metrics}
\end{equation}
Successful attacks in Eq.~\eqref{eq:metrics} change an initially correct prediction to an incorrect one.
Normalized partial ROC area (npAUC) integrates over FPR $[0,0.001]$ and divides by 0.001.

\subsection{Comparison with Existing Detectors}
\label{sec:comparators}
\label{sec:heldout-results}
To assess low-FPR detection and attack transfer, \cref{tab:latest-independent} compares eight frozen detectors across five attack configurations, including AutoAttack~\cite{croce2020reliable}.
Each detector uses one normal-data threshold calibration across all five attacks.
On PGD-40, {\codename} improves TPR over TTP by 4.40 percentage points and reduces false alarms by 69.7\%, from 89 to 27.
Its 0.027\% observed FPR has a one-sided 95\% upper bound of 0.0372\%.
Complete integer counts and FPR upper bounds appear in the appendix.

The largest TTP gap occurs on APGD-DLR, where {\codename} gains 40.40 points and reduces missed attacks by 98.7\%, from 1,251 to 16.
This attack is also the hardest for {\codename}, whose TPR remains within 99.48--100.00\% across the five configurations.
Because attacks and eligibility masks are shared, the gaps measure separation on the same inputs.
Maintaining detection with one threshold indicates transfer beyond the PGD training objective, while the lower normal-alarm count shows that this gain accompanies a more selective alarm set.
Comparisons on fixed 3,000-source and 256-source subsets appear in the appendix.
\begin{table}[!htbp]
\caption{Frozen detectors on 5,000 held-out ImageNet sources ($3/255$). Rates in \%; target FPR 0.1\%. Normal calibration and testing use 50,000 and 100,000 images, respectively. Attack columns report TPR. AA denotes AutoAttack-standard.}
\label{tab:latest-independent}
\centering\footnotesize\setlength{\tabcolsep}{2pt}
\begin{tabular}{lrrrrrr}
\toprule
Method & FPR$\downarrow$ & PGD-40 & PGD-100 & \shortstack{APGD\\CE} & \shortstack{APGD\\DLR} & AA\\
\midrule
Defend-CLIP & 0.053 & 13.80 & 10.80 & 0.10 & 53.91 & 0.26\\
GeoLID & 0.096 & 22.31 & 26.94 & 27.07 & 12.59 & 26.65\\
ResidualDiag & 0.080 & 19.37 & 27.14 & 40.44 & 4.78 & 40.90\\
ResidualMSE & 0.084 & 18.30 & 25.44 & 38.52 & 4.68 & 39.27\\
Feature Squeezing & 0.079 & 21.46 & 29.26 & 35.75 & 34.77 & 36.56\\
CSR & 0.099 & 76.55 & 93.64 & 98.79 & 18.55 & 98.99\\
TTP & 0.089 & 95.56 & 98.76 & 99.77 & 59.08 & 99.64\\
{\codename} & \textbf{0.027} & \textbf{99.97} & \textbf{99.97} & \textbf{100.00} & \textbf{99.48} & \textbf{100.00}\\
\bottomrule
\end{tabular}
\end{table}

\begin{figure}[t]
\centering
\includegraphics[width=0.9\columnwidth]{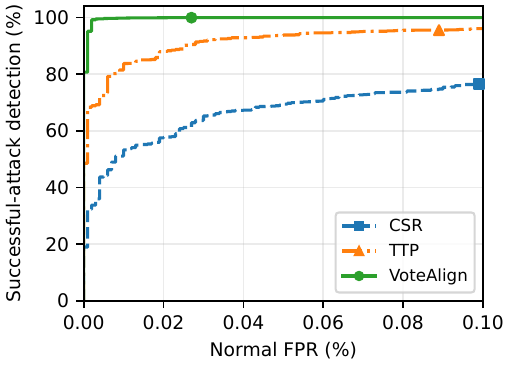}
\caption{Diagnostic ROC on held-out ImageNet PGD-40 ($3/255$), using 3,066 attacks and 100,000 test normals.
  Markers indicate calibrated deployment points.
  {\codename}'s curve varies both branch boundaries.}
\Description{CSR, TTP, and VoteAlign diagnostic ROC curves over observed FPR from zero to 0.1 percent.
  VoteAlign reaches 100 percent conditional detection by the 0.05 percent FPR cap; TTP and CSR are lower.
  Distinct marker shapes indicate each method's primary deployed operating point.
  Threshold scans of VoteAlign move both the mean and SVM boundaries.}
\label{fig:low-fpr-roc}
\end{figure}

To compare score ranking, \cref{fig:low-fpr-roc} scans the fixed PGD-40 scores.
{\codename} reaches 99.93\% TPR at an observed FPR cap of 0.02\%.
This supports low-FPR score separation, although scanning the curve moves both branch boundaries and the marked deployment point keeps the SVM boundary at zero.
\textbf{{\codename} combines the highest detection and lowest observed FPR across the five tested attacks.}

\subsection{Backbone and Domain Transfer}
\label{sec:cross-settings}
We assess backbone adaptation and domain transfer using PGD-40 on CLIP ViT-B/16 and three target datasets (\cref{tab:backbone-domains}).
B/16 uses separately fitted heads and an SVM on the same held-out identities.
Domain tests freeze the B/32 detector and score normalization.
Each domain has 2,000 test normals and 256 attack sources.
At a shared 0.5\% FPR target, Source uses 50,000 held-out ImageNet calibration normals and Target uses 3,000 target-domain normals.

On B/16, {\codename} matches TTP's complete detection and reduces false alarms by 88.1\%, from 84 to 10.
Across the three domains, it detects every eligible attack under both calibration policies.
On DTD, the five-point TPR gain over TTP under Target accompanies three additional false alarms.
Thus, complete detection can coexist with a domain-specific false-alarm cost.
Unchanged {\codename} detection and alarm counts between Source and Target indicate that recalibration adds no observed benefit on these samples.
Zero alarms on Food101 and EuroSAT describe their finite test sets.
\textbf{{\codename} extends effectively to different visual encoder architectures and classification tasks across datasets, generally improving detection rates and reducing false-positive rates relative to the baselines.}
\begin{table*}[t]
\caption{PGD-40 detection ($3/255$) in \%. B/16 uses 5,000 held-out sources, 50,000 independent calibration normals, and 100,000 test normals at target FPR 0.1\%. Domain transfer uses frozen B/32 detectors, 256 sources, 3,000 target-calibration normals, and 2,000 test normals per domain. Source uses 50,000 ImageNet normals and Target uses target-domain normals, both at target FPR 0.5\%. N/A denotes unfitted detectors.}
\label{tab:backbone-domains}
\centering\small\setlength{\tabcolsep}{3pt}
\begin{tabular}{llrrrrrrrr}\toprule
\multirow{2}{*}{Method} & \multirow{2}{*}{\shortstack{Domain\\calibration}} & \multicolumn{2}{c}{ImageNet (B/16)} & \multicolumn{2}{c}{Food101 (B/32)} & \multicolumn{2}{c}{DTD (B/32)} & \multicolumn{2}{c}{EuroSAT (B/32)}\\
\cmidrule(lr){3-4}\cmidrule(lr){5-6}\cmidrule(lr){7-8}\cmidrule(lr){9-10}
 & & TPR$\uparrow$ & FPR$\downarrow$ & TPR$\uparrow$ & FPR$\downarrow$ & TPR$\uparrow$ & FPR$\downarrow$ & TPR$\uparrow$ & FPR$\downarrow$\\\midrule
\multirow{2}{*}{Defend-CLIP} & Source & \multirow{2}{*}{0.64} & \multirow{2}{*}{0.070} & 35.32 & \textbf{0.000} & 56.00 & 0.550 & 8.91 & \textbf{0.000}\\
 & Target &  &  & 89.05 & 0.100 & 58.00 & 0.600 & 52.48 & 0.450\\
\midrule
\multirow{2}{*}{GeoLID} & Source & \multirow{2}{*}{27.89} & \multirow{2}{*}{0.082} & 74.63 & 2.000 & 57.00 & 0.900 & 84.16 & 56.950\\
 & Target &  &  & 44.78 & 0.200 & 47.00 & 0.550 & 28.71 & 0.200\\
\midrule
\multirow{2}{*}{ResidualDiag} & Source & \multirow{2}{*}{N/A} & \multirow{2}{*}{N/A} & 49.75 & 3.950 & 57.00 & 6.600 & 93.07 & 34.900\\
 & Target &  &  & 9.45 & 0.200 & 1.00 & 0.350 & 31.68 & 0.300\\
\midrule
\multirow{2}{*}{ResidualMSE} & Source & \multirow{2}{*}{N/A} & \multirow{2}{*}{N/A} & 47.76 & 3.150 & 57.00 & 6.500 & 91.09 & 33.000\\
 & Target &  &  & 9.45 & 0.150 & 1.00 & 0.350 & 26.73 & 0.250\\
\midrule
\multirow{2}{*}{Feature Squeezing} & Source & \multirow{2}{*}{38.18} & \multirow{2}{*}{0.084} & 7.46 & 0.650 & 50.00 & \textbf{0.150} & 38.61 & 0.450\\
 & Target &  &  & 5.47 & 0.150 & 54.00 & 0.350 & 34.65 & 0.150\\
\midrule
\multirow{2}{*}{CSR} & Source & \multirow{2}{*}{99.30} & \multirow{2}{*}{0.065} & 92.54 & 0.150 & 78.00 & 0.350 & 89.11 & 0.050\\
 & Target &  &  & 95.02 & 0.200 & 72.00 & 0.200 & 97.03 & 0.200\\
\midrule
\multirow{2}{*}{TTP} & Source & \multirow{2}{*}{\textbf{100.00}} & \multirow{2}{*}{0.084} & \textbf{100.00} & 0.150 & 95.00 & \textbf{0.150} & 99.01 & 0.050\\
 & Target &  &  & \textbf{100.00} & 0.400 & 95.00 & \textbf{0.100} & 99.01 & 0.050\\
\midrule
\multirow{2}{*}{{\codename}} & Source & \multirow{2}{*}{\textbf{100.00}} & \multirow{2}{*}{\textbf{0.010}} & \textbf{100.00} & \textbf{0.000} & \textbf{100.00} & 0.250 & \textbf{100.00} & \textbf{0.000}\\
 & Target &  &  & \textbf{100.00} & \textbf{0.000} & \textbf{100.00} & 0.250 & \textbf{100.00} & \textbf{0.000}\\
\bottomrule\end{tabular}
\end{table*}

\subsection{Component Studies}
\label{sec:controlled-studies}
We assess branch and head design on the shared ImageNet PGD-40 cohort.
Head-composition and ablation studies use five head seeds sharing one fixed SVM, with mean $\pm$ sample standard deviation (SD) reported across seeds, as detailed in the appendix.

\paragraph{Contribution of each branch.}
\label{sec:confirmation-attribution}
To assess SVM confirmation, \cref{tab:mean-attribution} compares the mean, SVM, and joint rules across ten complete fits.
Each fit retrains all heads and the SVM on paired bootstrap samples from one training pool and uses development data for model selection.
Confirmation reduces mean-only false alarms by 61.9\%, with mean misses increasing from 1.6 to 1.8 per fit.
This small detection change indicates that the gate removes many normal-tail alarms while retaining most attacks flagged by the mean.
Against the zero-threshold SVM, false alarms fall by 38.5\%, with 1.3 additional misses per fit.
\begin{table}[!htbp]
\caption{Branch comparison on held-out ImageNet PGD-40 ($3/255$), with 3,066 successful attacks and 100,000 test normals. All rules use the same ten complete fits, each with its own heads and SVM. The mean threshold uses all 50,000 normal calibration images at target FPR 0.1\%. The SVM boundary is zero. Results show mean $\pm$ sample SD across fits.}
\label{tab:mean-attribution}
\centering\small\setlength{\tabcolsep}{1.4pt}
\begin{tabular}{lrrr}
\toprule
Rule & npAUC$\uparrow$ & TPR (\%)$\uparrow$ & FP$\downarrow$\\
\midrule
Mean only & 0.9959$\pm$0.0007 & 99.95$\pm$0.02 & 77.1$\pm$7.5\\
SVM alone ($d>0$) & 0.9946$\pm$0.0010 & \textbf{99.98$\pm$0.03} & 47.8$\pm$9.5\\
{\codename} & \textbf{0.9966$\pm$0.0004} & 99.94$\pm$0.03 & \textbf{29.4$\pm$5.4}\\
\bottomrule
\end{tabular}
\end{table}

At the same nominal FPR target ($0.1\%$), we calibrate the standalone SVM on the shared normal split.
The joint rule reduces false alarms by 62.3\%, with 1.6 additional misses per fit.

We also maximize TPR under a post-hoc cap of 30 false alarms per fit, keeping the joint rule's SVM gate fixed.
At the best TPR under this post-hoc cap, mean misses are 3.0 for the mean alone, 1.7 for the joint rule, and 0.9 for the standalone SVM.
Paired differences across fits are reported in the appendix.
\textbf{SVM confirmation substantially reduces the mean branch's false positives with only a slight decrease in TPR.}

\paragraph{Head composition and MLP controls.}
To test head design and capacity, \cref{tab:member-capacity} compares individual heads and MLP controls.
Each base rule receives its own NP threshold and the same SVM gate.
The controls are a $2304\rightarrow86\rightarrow1$ MLP and a mean of three $2304\rightarrow29\rightarrow1$ MLPs.
Both read blocks 4/8/12 and use the same losses.
The full model uses three times as many head optimizer updates as the single MLP.

The full model reduces mean missed attacks by 40.0\% relative to the single MLP and 45.5\% relative to the homogeneous mean, with npAUC gains of 0.00298 and 0.00271.
It adds one false alarm relative to the single MLP and matches the homogeneous mean's alarm count.
The individual-head results expose the tradeoff behind fusion.
$H_3$ raises the fewest alarms but misses more attacks, while the full model improves detection over each head.
\begin{table}[!htbp]
\caption{Individual heads and MLP controls on ImageNet PGD-40 ($3/255$), with 3,066 successful attacks, 100,000 test normals, and all 50,000 held-out calibration normals. Mean $\pm$ sample SD over five head seeds sharing one fixed SVM. All rows include the SVM gate.}
\label{tab:member-capacity}
\centering\small\setlength{\tabcolsep}{3pt}
\resizebox{\columnwidth}{!}{%
\begin{tabular}{lrrr}
\toprule
Configuration & npAUC$\uparrow$ & TPR (\%)$\uparrow$ & FP$\downarrow$\\
\midrule
\multicolumn{4}{l}{\textit{Individual heads}}\\
$H_1$ & 0.9941$\pm$0.0007 & 99.92$\pm$0.03 & 26.4$\pm$1.1\\
$H_2$ & 0.9947$\pm$0.0005 & 99.92$\pm$0.04 & 30.6$\pm$3.4\\
$H_3$ & 0.9954$\pm$0.0003 & 98.71$\pm$0.13 & \textbf{8.0$\pm$0.0}\\
\midrule
\multicolumn{4}{l}{\textit{MLP controls}}\\
Single MLP & 0.9940$\pm$0.0004 & 99.93$\pm$0.02 & 26.2$\pm$1.1\\
Homogeneous mean & 0.9943$\pm$0.0003 & 99.93$\pm$0.01 & 27.2$\pm$1.6\\
\midrule
\multicolumn{4}{l}{\textit{Full model}}\\
{\codename} & \textbf{0.9970$\pm$0.0005} & \textbf{99.96$\pm$0.01} & 27.2$\pm$1.3\\
\bottomrule
\end{tabular}%
}
\end{table}

Feature-depth and training-loss ablations are reported in the appendix.
\textbf{Heterogeneous heads improve observed detection and ranking over the MLP controls, with a small false-alarm cost.}

\paragraph{Training and calibration set sizes.}
\label{sec:sensitivity}
To measure how much data the same detector needs, we vary the number of training pairs and calibration images as detailed in the appendix.
Increasing training pairs from 256 to 8,000 reduces mean missed attacks by 90.3\%, from 10.3 to 1.0, with little further gain beyond 4,000 pairs.
Detection improves across these training budgets, while mean FPR varies nonmonotonically between 0.012\% and 0.031\%.

For fixed models, increasing calibration images from 3,000 to 10,000 narrows the TPR range by 92.3\%.
From 3,000 to 20,000 calibration images, the observed FPR range also narrows from 0.007--0.036\% to 0.024--0.031\%, while mean FPR rises from 0.0206\% to 0.0261\%.
This narrower range indicates reduced sensitivity to calibration-sample selection for fixed models.
These studies separate sensitivity to detector fitting from threshold sensitivity for fixed models.

\Needspace{3\baselineskip}
Complete curves are shown in the appendix.
Additional results for normal transformations and operating conditions are reported in the appendix.
\par\Needspace{5\baselineskip}\noindent
\textbf{Within the tested ranges, more training pairs improve detection and reduce misses, with diminishing gains beyond 4,000 pairs. Larger normal calibration pools reduce variability in TPR and FPR across subset draws, stabilizing the operating point.}

\Needspace{13\baselineskip}
\paragraph{Evaluation scope and calibration assumptions.}
\label{sec:discussion}
High detection across attacks supports transfer under frozen calibration, while the joint rule can reject attacks accepted by a standalone branch, as detailed in the appendix.
Transformation-dependent recalibration effects further show that calibration gains depend on the normal distribution, as shown in the appendix.
The NP guarantee requires independent i.i.d.\ calibration normals from the target distribution (\cref{sec:vote-calibration}).
\textbf{The evidence supports low-FPR detection under the tested classifier-targeted attacks, with repeated-fit estimates conditional on the shared cohort.}

\paragraph{Computational cost.}
To assess inference overhead, encoding and detection times on prepared held-out inputs using an RTX A6000 and float32 are reported in the appendix.
{\codename} adds 2.62 ms per image over CLIP classification, reducing time by 26.7\% relative to CSR and 37.5\% relative to TTP.
Both branches reuse classification features, consistent with the modest measured overhead of evaluating several scores in one encoder pass.
ResidualMSE remains 8.5\% faster than {\codename}, placing the joint detector above the lightest residual control in latency.
\textbf{{\codename} reduces false-positive rates and improves detection rates with only a modest increase in inference latency.}

\FloatBarrier
\section{Conclusion}
We propose {\codename} to reduce false alarms while retaining high detection rates in adversarial detection.
It combines three independently trained heads with classification and head-specific ranking objectives, aligned mean fusion, fixed SVM confirmation, and normal-data Neyman--Pearson calibration.
Both branches reuse classification features.
Our analysis characterizes ranking-loss sensitivity and confirmation tradeoffs, and provides finite-sample FPR and TPR guarantees under independent normal calibration and attack audits.
At a common nominal FPR target, {\codename} detects 99.97\% of successful held-out PGD-40 attacks versus 95.56\% for TTP, with 27 versus 89 false alarms per 100,000 normals.
Across five attacks, detection remains 99.48--100.00\% at 0.027\% observed FPR after one normal calibration.
The method provides a new baseline for low-FPR adversarial detection, with joint evaluation of detection, false alarms, and inference cost.

\label{sec:main-text-end}
\clearpage
\hypersetup{pageanchor=false}
\onecolumn
\thispagestyle{empty}
\thispdfpagelabel{LLM Usage Statement}
\section*{LLM Usage Statement}

The authors take full responsibility for all submitted content, regardless of any LLM or AI-tool use.
The authors use OpenAI (ChatGPT / GPT) to obtain reviewer-style feedback on the manuscript.

\paragraph{Writing and editing (Generate).}
The tool generates draft prose and revisions for language, structure, and clarity.
The authors review, edit, or approve the generated text.

\paragraph{Research and preparation (Assist).}
The tool helps edit or refine author-produced work on related work and citations, code and implementation, experiment design and execution, result analysis, figures and other visual materials, data preparation, and research ideas, hypotheses, claims, and conclusions.

\twocolumn
\addtocounter{page}{-1}
\hypersetup{pageanchor=true}
{\small
\bibliographystyle{ieeenat_fullname}
\bibliography{references}
}
\end{document}